\documentclass[11pt]{article}
\usepackage[english]{babel}
\usepackage{amsmath,amsthm}
\usepackage{amsfonts}
\usepackage{hyperref}
\usepackage{graphicx}
\usepackage{enumerate}
\usepackage[round]{natbib}

\begin{document}

\title{Analysis of Microarray Data using Artificial Intelligence Based Techniques}

\author{Khalid Raza \\
        Department of Computer Science \\
        Jamia Millia Islamia (Central University), New Delhi, India \\
        Email: kraza@jmi.ac.in
}

\date{June 30, 2015}
\maketitle
\begin{abstract}
  Microarray is one of the essential technologies used by the biologist to measure genome-wide expression levels of genes in a particular organism under some particular conditions or stimuli. As microarrays technologies have become more prevalent, the challenges of analyzing these data for getting better insight about biological processes have essentially increased. Due to availability of artificial intelligence based sophisticated computational techniques, such as artificial neural networks, fuzzy logic, genetic algorithms, and many other nature-inspired algorithms, it is possible to analyse microarray gene expression data in more better way. Here, we reviewed artificial intelligence based techniques for the analysis of microarray gene expression data. Further, challenges in the field and future work direction have also been suggested.
\end{abstract}

\section{Introduction}

The bioinformatics is an interdisciplinary area of study where one of the objectives is to deal with the analysis and interpretation of large sets of data generated from various large-scale biological experiments. The example of one such large-scale biological experiment is measuring the expression levels of tens of thousands of genes simultaneously under some environmental condition. Microarray is one of the essential technologies used by the biologist to measure genome-wide expression levels of genes in a particular organism. As microarrays technologies have become more prevalent, the challenges associated with collecting, managing, and analyzing the data from each experiment have essentially increased. Robust laboratory protocols, improved understanding of the complex experimental design and falling prices of commercial platforms, all these have combined to drive the field to more complex experiments, generating huge amounts of data  \citep{brazma2000}.

	With the help of measured transcription levels of genes under different biological conditions (e.g. at various developmental stages and in different tissues), biologists are able to develop gene expression profiles that differentiate the functionality of each gene in the genome. The gene expression profiles are organized in the form of a matrix, where rows represents genes, columns represents samples/replicas, and each cell of the matrix contains a numeric value representing the expression level of a gene in a particular sample. Generally, such as table is called gene expression matrix. If over expression of certain genes is correlated with a certain disease then researchers can discover what are other conditions affecting expression-level of these genes. Also, what are the other set of genes having similar expression profiles pattern. Hence, suitable compounds (potential drugs) can investigated that can lower the expression level of these overexpressed genes \citep{babu2004}.

Many sophisticated statistical and computational tools have been developed to help biologists for the analysis of gene expression data and to identify novel targets from their experimental data \citep{deng2009, debouck1999}. Among these techniques, clustering and statistical methods are most commonly used data analysis methods. Clustering generally groups the gene expression data with similar expression pattern, i.e. co-expressed genes. However, clustering approach suffers from several drawbacks \citep{bassett1999}. The statistical methods help to analysis gene expression data and infer relationships between genes. However, it fails to provide complex regulatory relations among genes.
	
The chapter is organized as follows. Section 2 describes the background of Microarray experiments and data generation. Section 3 covers the applications of Microarrays and Section 4 describes artificial intelligence based techniques, and reviews its application in the analysis of Microarray data. Section 5 summarizes the chapter and presents research challenges and future work directions.

\section{Microarray Technology}

With the help of Microarray technology, one can measure the expression level of all genes in a genome simultaneously. By measuring and comparing the expression level of genes in an unhealthy versus healthy cell, it would be possible to identify genes which are responsible for various diseases. Due to unprecedented amount of large biological data generated out of microarray experiments, research’s focus has shifted from the generation of data to the analysis and presentation of data in the most efficient manner \citep{hood2003, kitano2002a, kitano2002b}. With the help of these technologies, researchers can find out answer to some of the challenging questions like;

\begin{enumerate}[(i)]
  \item What are the functions of different genes?
  \item In what cellular processes do these genes participate?
  \item How genes are regulated?
  \item How genes and its products (proteins) do interact, and what are these interaction networks?
  \item How expression level of genes differs in different cell types and states?
  \item How expressions of genes are affected by various disease or drug treatments?
\end{enumerate}

Microarrays are frequently used in biomedical research to tackle a number of problems, including classification of tumors, or gene expression response to different stress conditions. A central and frequently asked question in microarray is the identification of differentially expressed genes (DEGs). The DEGs are those genes whose expression levels are associated with a response or covariate of interest \citep{dudoit2002}. The covariates can be either polytomous (for instance, treatment/control status, cell type, drug type) or continuous (for instance, drug doses), and the responses can be, for instance, censored survival times or any other clinical outcomes \citep{dudoit2002, lee2012}. Scientists from different disciplines such as biology, statistics, computer science, mathematics, bioinformatics, etc. are working in this area to identify some new insight from DNA microarray data such as identification of differentially expressed genes, classification between cancerous and non-cancerous genes, identification of potential genes for drug target, identification of gene function, and so on. In the following section, various steps involved in Microarray experiments have been discussed.

\subsection{ Experimental Setup}
\begin{figure}[h]
  \centering
  \includegraphics[width=11cm]{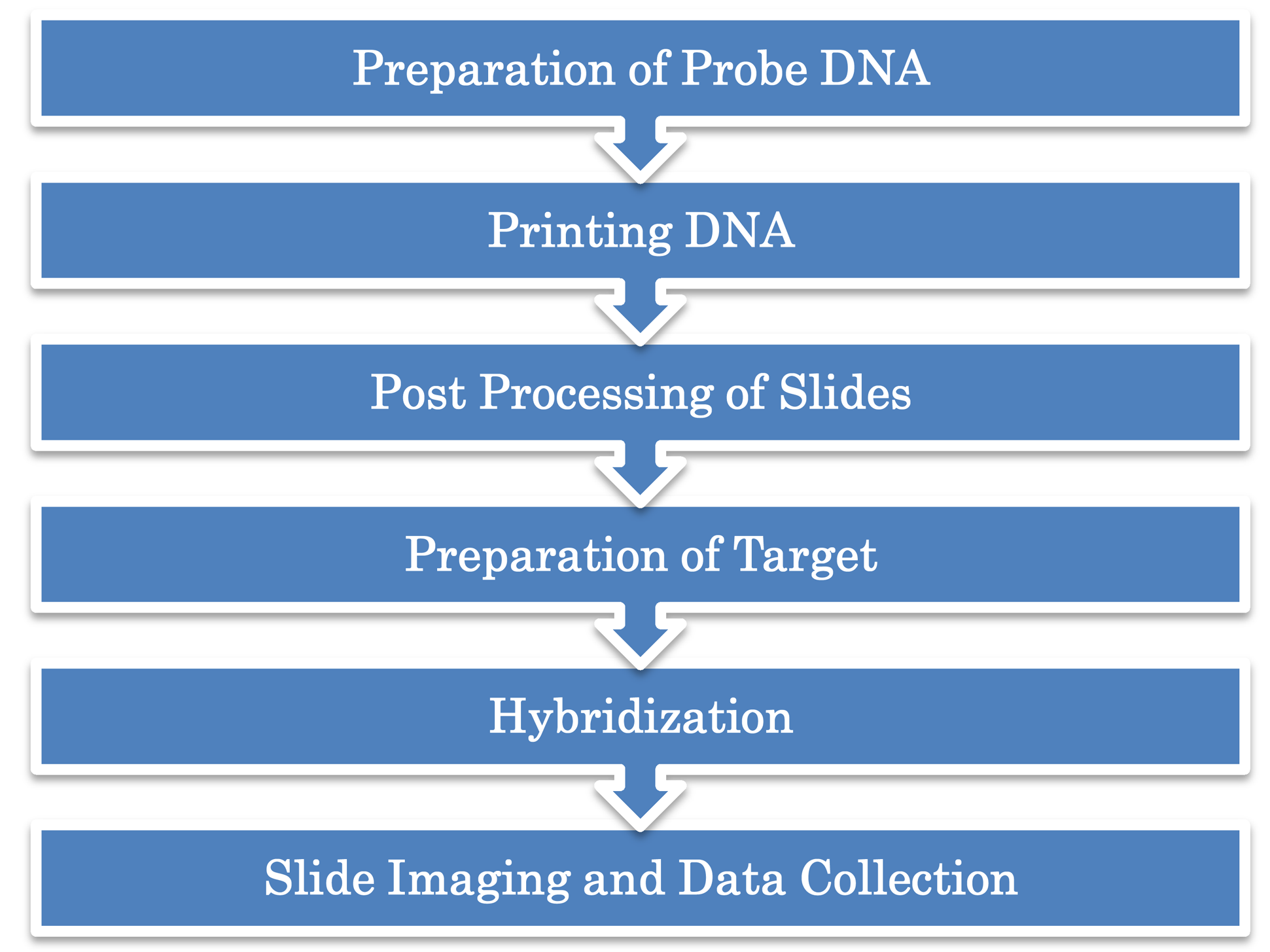}
  \caption{Step involved in microarray experiments}\label{f1}
\end{figure}

Basically microarray is a solid base having grid of spots where genetic material of known sequence is arranged systematically. It is mostly made up of glass on which single stranded DNA molecules are attached at fixed positions. The size of the arrays can vary from microscope slide to square silicon chips. On an array, there can be thousands of spots and each spots contain number of identical DNA molecules. The microarray fabrication can be done in two ways: i) cDNA and ii) oligonucleotide. The steps involved in microarray experiments are shown in Fig. \ref{f1}.

\subsubsection{Preparation of probe DNA}
For the study of large-scale expression, a specific DNA sequence is needed for all genes whose expression values are to be measured. The selection of probe is done on the basis of resources available for obtaining the representation of the genes under studies. The simplest way is to amplify every known ORF in the genome and use it as a probe. PCR is used for the amplification purpose and allows multiplication of DNA fragments by millions in just few hours. ESTs may be used to identify distinct mRNA transcripts.

\subsubsection{Printing array}
In cDNA array, arrays are mostly printed on Poly-L-lysine coated glass microscope slide using arraying robot. During the arraying operation, a large number of slides are placed on and secured to a platter. The samples of DNA are placed in microliter plates on the stand. The reservoir slot of each tip is filled with ~1µ liter of DNA solutions. The tips are then lightly tapped at identical positions on each slide leaving a small drop of DNA solution on the poly-L-lysine coated slide. In Oligonucleotide array, oligos are printed at the spots instead of cDNA. Same robotics can be applied to manufacture both types of arrays. However, the preparation of oligonucleotide array is quite different. During fabrication of array, the probes are synthesized on the chip using photolithography.

\subsubsection{Post processing of slides}
This step consists of Rehydration and Blocking. The spots on the microarray are rehydrated to distribute DNA more evenly. In the blocking process, free reactive groups on the slide surface are modified to minimize their ability to bind to labeled target DNA. If these groups are not blocked, the labeled DNA target can bind to the surface of the slide.

\subsubsection{Preparation of target}
In this step, isolate mRNA from the samples and purify it. Since, mRNA degrades very fast; hence it is reverse-transcribed into more stable cDNA.

\subsubsection{Hybridization}
In hybridization process, a single stranded DNA molecule is bound to another single strand DNA molecule with a precisely matching sequence. After hybridization process, the microarray properly washed to eliminate any excess labeled sample and finally dried using a centrifuge. Sometimes two types of target mRNA samples are simultaneously hybridized on the array, called two channel microarray experiment. In that case, two types of molecules are added to targets and uses fluorescent dyes like Cy3 and Cy5, which can be separated spectrally. The Cy3 is green and Cy5 is red when excited by laser light at specific wavelength.

\subsubsection{Slide imaging}
\begin{figure}[h]
  \centering
  \includegraphics[width=13cm]{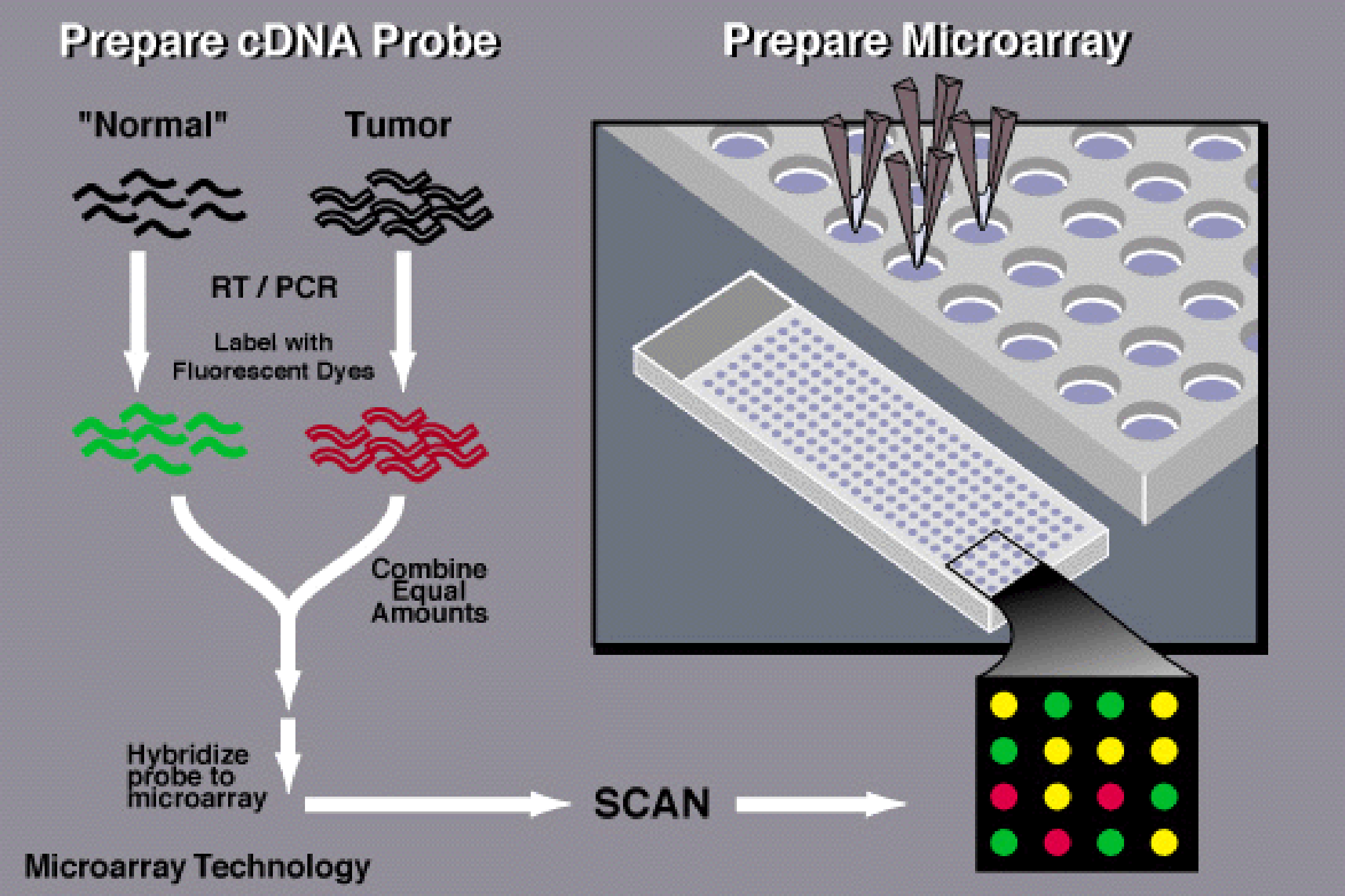}
  \caption{Preparation of cDNA probe and microarray}\label{f2}
\end{figure}

Under this step, the microarray is scanned to measure the fluorescent signal emitted at every spot that determine the amount of labeled sample bound to each spot. The laser scanning confocal microscope is used for this purpose. For single channel array, the array is scanned once but for two-channel experiment, it is scanned in two phases. In the obtained image, the intensity of each spot is proportional to the amount of mRNA from the sample, matching cDNA sequence of given pot. A gene expressed in a sample labeled with red dye and not expressed in the other sample will produce a red spot and vice versa. A gene expressed in both the samples will produced equal amount of red and green intensities and a spot is given yellow (Fig. \ref{f2}).

\subsection{Quantification of Images}
The images generated by scanner of microarray are the raw data. There is one image per array for a single channel microarray, while there are two images per array for two channel microarrays. The image intensity is scanned by detector at a high spatial resolution, where every probe spots are represented by several pixels. The intensity values of each probe are identified and these intensities are quantified to numeric values. Image quantification process involves following steps:
\begin{enumerate}[(i)]
  \item Identification of position of sports on the array
  \item For every spot on the array, pixel identification on the image
  \item For every spot on the array, identifying pixels so that it may be used for background calculation
  \item Computation of numeric value for the intensity of the spot, intensity of background image and quality control information.
\end{enumerate}

There are several methods for segmentation and quantification which are available in software packages but they differ in their robustness. To a microarray project, quantification of image involves the transition of workflow from wet-lab procedure to computational (dry-lab). During the computation of numeric information at the microarray spot, image processing software provides a number of measures such as mean, median and standard deviation of signal and background, along with diameter and number of pixels. Among these measures, the most important measure is hybridization intensity for each spot that can be either mean or median of the pixel intensities. The second important information is signal standard deviation that helps in computation of coefficient of variation for spot as well as for the background. Once the spotted image and other statistics are computed then it is suggested that quality of the array and individual spots on the array are assessed because sometimes array may have a few spots as defected.

\begin{figure}[h]
  \centering
  \includegraphics[width=8cm]{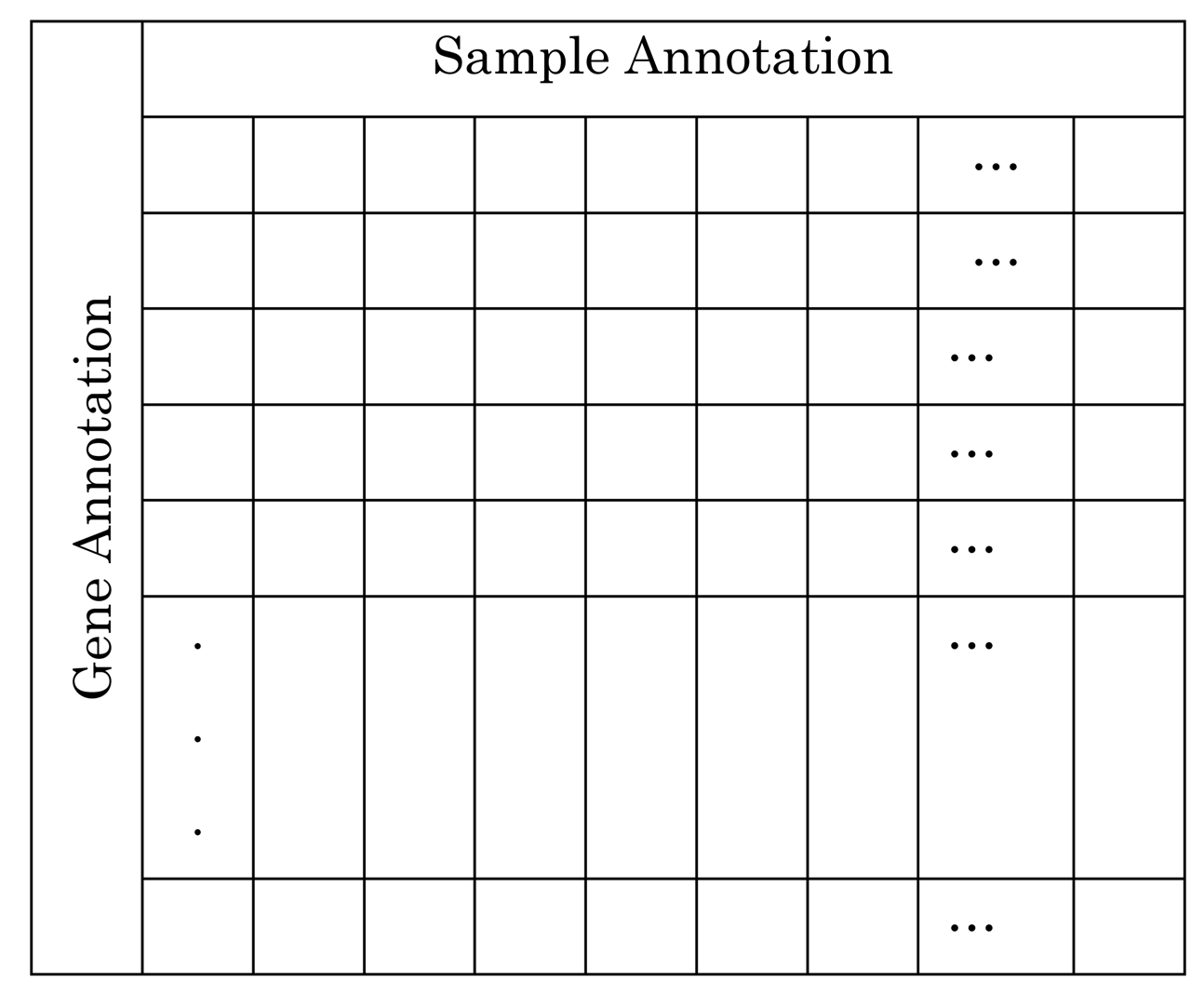}
  \caption{Preparation of cDNA probe and microarray}\label{f3}
\end{figure}

The obtained microarray gene expression data is presented in a tabular form, called gene expression matrix, as shown in Fig. \ref{f3}. The first column specifies the identity of genes and first rows represents \emph{samples} (replicas), \emph{condition} or \emph{time-series} observations. The element of the matrix gives expression values of gens under various conditions or samples.

\subsection{Data Preprocessing and Normalization}
Once the spotted images have been quantified to generate datasets, these datasets should be preprocessed before its analysis and interpretation. In this step, meaningful characteristics are extracted or enhanced and prepare the dataset for its analysis and interpretation. In data preprocessing step, generally two issues are addressed: i) to adjust background intensities, and ii) to transform data into scale suitable for analysis and interpretation. A simple example of preprocessing microarray data is taking the Log of the raw intensity values. The main purpose of normalization is to ensure that variation in the expression values are because of biological differences between the mRNA samples and not because of experimental artifacts.

The adjustment of background intensities are needed because despite of washing done after hybridization in microarrays, there are chances of genes annealing in the background of the spot and during scanning time, it may give rise to background intensity. Another issue in the gene expression data which need to be addressed is the difference between the data generated by two microarray technologies (cDNA and oligonucleotide microarrays). The cDNA reports differences in gene expression, while oligonucleotide microarray report absolute expression values \citep{butte2002}. Hence, same normalization techniques may not be applied to these different microarray technologies. In a given experiment, most genes do not changes their expression levels and if equal numbers of genes are upregulated and downregulated, then differential expression measurements might found to be normally distributed.

\subsubsection{Normalization for single channel experiment}
Suppose that one need to find out differentially expressed genes (DEGs) under various experimental conditions, such as normal sample against cancerous sample, control tissue versus treatment, etc. in a single channel experiment. It is generally expected that gene expression values in both the conditions are more or less similar but it is seldom found in the reality. This variation is because of many factors including different arrays are used for each samples. Hence, it is natural that one would expect distribution of expression values would more or less similar. Therefore, it is necessary to remove variation between arrays and the methods to remove variation among arrays, and is called array normalization methods. There are several methods to make empirical distribution of expression values over all arrays. Some of the methods are:

\begin{enumerate} [(i)]
  \item Normalization by mean
  \item Median or Q2 normalization
  \item Q3 normalization
  \item Quantile normalization
\end{enumerate}

In normalization by mean method, the expression values are transformed so that that mean of all the arrays is same. Median method transforms the expression values so that all arrays have median same as that of some reference array. The Q3 method is defined as similar to Q2. In Q3, third quartiles of the arrays are calculated to the third quartile of the mock array. Quantile normalization method is an extension of Q2 and Q3 normalization. This method is based on transforming each array specific distributions of intensities so that they all have similar values of quantiles.

\subsubsection{Normalization for two channel experiment}
In a two channel microarray experiment, two different samples are labeled by two fluorescent dyes $R$ and $G$, hybridized on an array. The difference of intensities between two channels gives DEGs, provided that these variations are only because of biological functioning of the genes in different conditions. Here, to identify DEGs it is necessary to compare the intensity of $R$ and $G$. These methods are applied on the Log of the ratio $R/G$.

\section{Applications of Microarrays	}
Microarrays have been utilized in several biomedical problems including gene discovery, disease diagnosis, pharmacogenomics, and toxicology. For instance, microarrays can be used to identify disease genes by comparing expression patterns of genes in disease versus normal cells sample. Similarly, it can also be used to identify possible abnormal gene expression and abnormal interaction between genes for a disease.

For a majority of applications, microarrays address four broad categories of problems \citep{xu2008}:
\begin{enumerate}[(i)]
  \item Gene selection/gene filtering or identification of differentially expressed genes (DEGs)
  \item Finding natural groupings among genes, conditions or both (clustering)
  \item Patient classification using gene expression
  \item Finding regulatory relationships among given set of genes
\end{enumerate}

\subsection{Gene Selection or Identification of DEGs}
An important purpose for monitoring expression level of genes is to identify those genes which are differentially expressed across two kinds of tissue samples or samples observed under two different experimental conditions. Set of genes differentially expressed over two different samples, i.e., normal and cancerous tissue, are expected to give clues about cancer mechanism. A large variety of methods exists for finding differentially expressed genes and most of these methods are based on statistical techniques, such as fold-change \citep{schena1995}, t-test statistics \citep{peck2011, druaghici2003}, ANOVA \citep{kerr2000}, rank product \citep{breitling2004}, Significant Analysis of Microarray \citep{tusher2001}, Random Variance Model \citep{wright2003}, Limma \citep{smyth2004}, and so on. Review on the various methods can be found in: \citep{pan2002, jeffery2006}.

Fold change is one of the simplest ad-hoc methods often used in microarray analysis. A fold change is a measure that describes how much expression level of a gene changes over two different samples (conditions) or groups. To calculate a fold change, the average of expression values for each probes are calculated across the samples in each group, and then ratios of these average are taken. The levels of fold change are observed and genes under or above a thresholds are selected. For example, fold change below 0.5 is considered as down-regulated and fold change above 2.0 is considered as up-regulated. The Rank Products method is based on the statement that an experiment examining for n genes in m replicas, will have probability to be ranked first of $1/nm$, if the list values were totally random. Hence, it is improbable that single gene to have top position in all the given replicas, if given gene was not expressed differentially. Then, genes can be sorted based on likelihood of observing their rank product values \citep{jeffery2006}. The two-sample t-test is widely used parametric hypothesis testing method for the identification of DEGs. The t-statistics gives a probability value (p-value) for each gene. A small p-value indicates that genes are differentially expressed under the hypothesis that there is no differential expression, which is not true. The t-statistic is calculated as the difference in the means over the standard deviation. \citet{raza2012} proposed an anticlustering gene algorithm for the identification of genes as drug target, where they applied a combination of statistical techniques.

\subsection{Clustering Genes, Samples or Both}
Clustering is a means of analysing set of objects by grouping them into different clusters based on some similarity measures. Basically clustering is an unsupervised technique that groups the similar objects into clusters. Researchers can apply clustering techniques to cluster gene expression data. Hence, genes belonging to a particular cluster are supposed to share common properties.  If gene expression profiles (genes) are clustered, one may discover set of genes co-regulated in a certain samples. Similarly, gene expression can be grouped by clustering its samples. Sample clustering is done when it is needed to identify subgroups of certain condition (for instance, disease). The third means of grouping the gene expression data is to cluster both rows (genes) as well as columns (samples), which are known as co-clustering or bi-clustering. This kind of clustering helps us to find groups of genes associated with group of samples (patient). Some of the most popularly used clustering techniques are k-means, hierarchical, SOM, fuzzy c-means, non-Euclidean relational fuzzy c-means.

Using Microarray gene expression data, distance between two expressed genes can be computed so that it can be known that whether genes are interrelated or not, and placed in same cluster. Euclidean distance, Manhattan distance and Pearson correlation distance are some of the commonly applied distance measures. In \citet{raza2014}, four different clustering techniques viz., k-means, Hierarchical clustering, density based clustering and Euclidean method based clustering, have been applied on five different types of cancer gene expression data (lung cancer, prostate cancer, colon cancer, breast cancer and ovarian cancer). In all these five datasets, there are large numbers of genes compared to numbers of samples. Hence, for better learning on machine learning techniques and to avoid the “curse of dimensionality problem”, after data normalization, attribute reduction using t-test has been done at a significance level of 0.001. There is no single clustering algorithm that can work well in all the situations. Selection of a particular clustering approach depends on the problem at hand and the dataset under study.

\subsection{Patient Classification}
Gene expression data can be used to train a classifier so that it can recognize a given condition (e.g. class label such as normal or cancerous). The advantage of this kind of classification is that once a classifier is trained with gene expression profiles to recognize a patient class, then it can recognized a class of unknown patient for which the classifier has not been trained. There are several supervised techniques available for patient classification, such as Bayesian networks, M5 model tree, k-nearest neighbourhood, Random forest, neural networks, and support vector machines.

In \citet{razahasan2013}, authors have done a comparatively evaluation of various machine learning techniques for their accuracy in class prediction of prostate cancer based on Microarray dataset. As per their evaluation, Bayes Net gave the best accuracy for prostate cancer class prediction with an accuracy of 94.11\%. Bayes Net is followed by Navie Bayes with an accuracy of 91.17\%. The objective of evaluating various machine learning techniques is to come up with the best technique in terms of prediction accuracy and to reveal a good procedure for meaningful attribute reduction. A similar kind of process may be used to classify other types of cancers. One of the biggest challenges is to develop a single universal classifier which would be capable of classifying all types of cancer gene expression data into meaningful number of classes.

\subsection{Finding Regulatory Relationship among Genes}
A gene regulatory network (GRN) is a network of interaction among genes, where node represents genes and interconnection between them represents their regulatory relationship. Today, one of the most exciting problems in systems biology research is to decipher how the genome controls the development of complex biological system. Microarrays have been widely used to find out new (unknown) regulatory mechanism. The discovery of GRN using gene expression data is known as reverse-engineering of GRN. The GRNs help in identifying the interactions between genes and provide fruitful information about the functional role of individual genes in a cellular system. They also help in diagnosing various diseases including cancer.

In the last several decades, many computational methods have been proposed to discover complex regulatory interactions among genes based on microarray data. These techniques can be clubbed into different groups, such as Boolean networks \citep{liang1998, akutsu1999, shmulevich2002, martin2007, raza2013reconstruction, raza2013soft}, Bayesian networks \citep{friedman2000, husmeier2003}, Petri nets \citep{koch2005, remy2006}, linear and non-linear ordinary differential equations (ODEs) \citep{chen1999, tyson2002, de2008search}, machine learning approaches \citep{weaver1999, kim2000, vohradsky2001, keedwell2002, huang2003, tian2003, zhou2004, xu2004inference, hu2006, jung2007, xu2007modeling, xu2007inference, chiang2007, lee2008, datta2009, zhang2009, maraziotis2010, ghazikhani2011, liu2011, kentzoglanakis2012, noman2013}, etc.

For review of the modeling techniques and the subject, refer to \citep{de2002modeling, wei2004, schlitt2007, cho2007, karlebach2008, swain2010, sirbu2010, mitra2011, raza2013soft}.

\section{Artificial Intelligence Techniques and Microarray Analysis}
Artificial Intelligence (AI) is an interdisciplinary field of study where the goal is to create intelligence by a machine or a computer program. Most of the researchers defines AI as “the study and design of intelligent agents”, where an intelligent agent is a system that can perceive given environment to take actions and maximize its probability of being success. John McCarthy coined this term in 1955 and defined AI as “the science and engineering of making intelligent machine”. AI has a vast domain of research including reasoning, knowledge, learning, natural language processing, perception, etc. Most popularly used AI techniques for solving real-life problems, including bioinformatics problems such as analysis of microarray array data to extract fruitful knowledge, are statistical methods and computational intelligence. Among the computational intelligence, artificial neural networks, fuzzy systems, evolutionary computations and many statistical tools are mostly applied AI based approach.

In this section, few computational intelligence approaches and their applications in Microarray analysis have been briefly described.

\subsection{Artificial Neural Networks}
Artificial Neural Networks (ANNs) are massively parallel computing system which is inspired by biological system of neurons. It is collection of extremely large number of simple processing elements, called neurons, having many interconnections. These elements have inputs, which are multiplied by weights and then computed by a mathematical function, called activation function, regulating the activation of the neuron. A weight value $w_{ij}$ is assigned to each connection and hence the net input to the neuron is the weighted sum of its $n$ input signals $x_i$, $i=1, 2,.., n$. Each neuron has an activation function f (generally a sigmoidal function), which is used to compute the neuron’s current activation ai, and output function g (generally identity function) which is used to compute value $O_i$. By adjusting the weights of neurons, the desired output can be obtained from the inputs. The process of adjusting the synaptic weights is known as learning or training process. The most widely applied training algorithm is Backpropagation algorithm.

\begin{figure}[h]
  \centering
  \includegraphics[width=13cm]{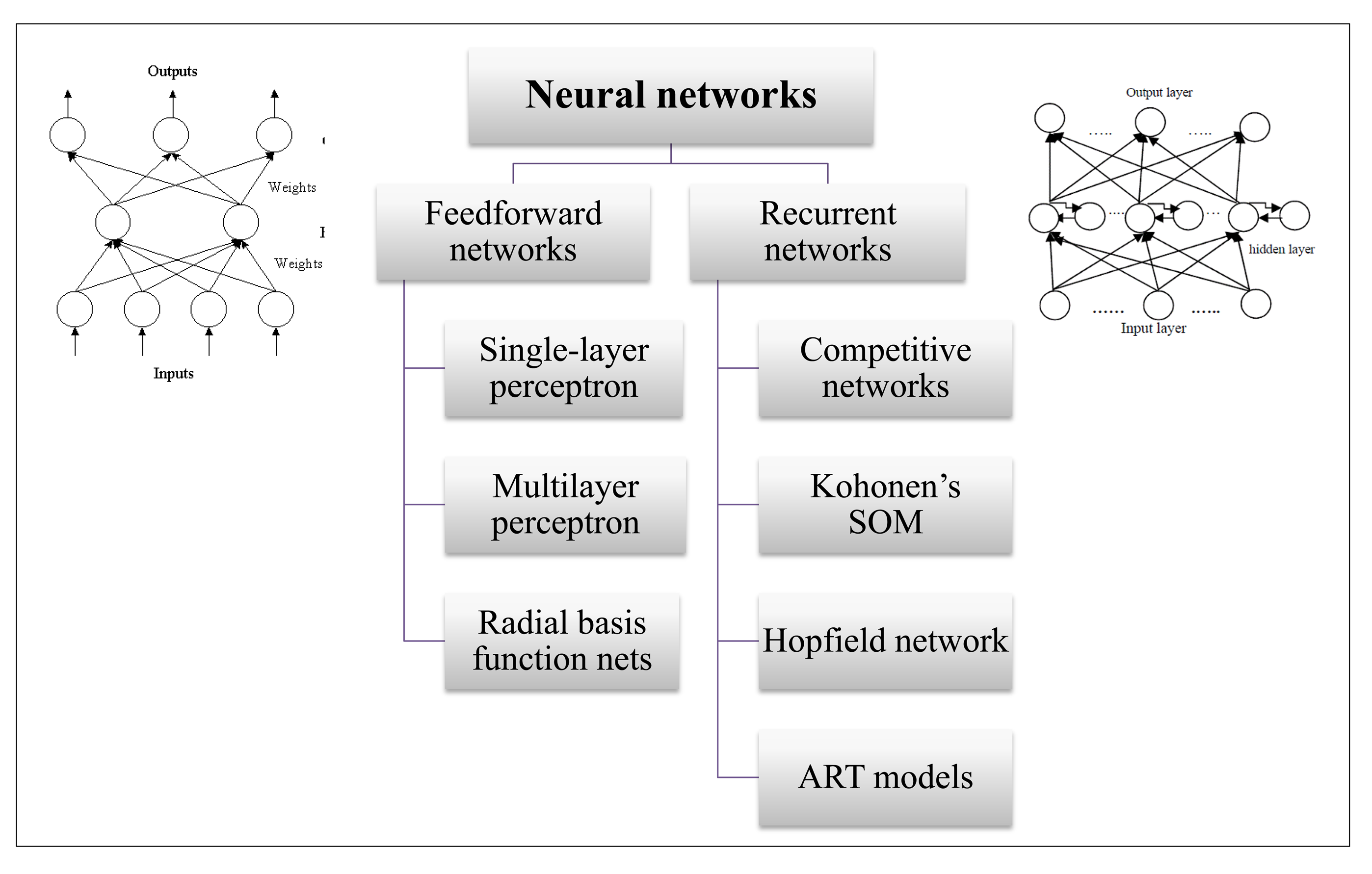}
  \caption{A taxonomy of feed-forward and recurrent network architectures}\label{f4}
\end{figure}

The first neural network model was proposed by McCulloch and Pitts in 1993 and since then hundreds of different models have been developed \citep{gershenson2003}. The differences in the neural networks might be in their architecture, activation function, and topology, training algorithm and accepted input and output values. On the basis of architecture (connection pattern), neural networks can be broadly clubbed into two groups \citep{jain1996}: i) feedforward networks, where there is no loop and ii) recurrent networks, which contain loops because of feedback connections. A typical network from each category is shown in Fig. \ref{f4}. Feedforward networks are memory-less, i.e., their output is independent of the previous network state but recurrent networks consider network feedback paths, which are with memory, i.e., the current output is dependent on the previous state of the network. Different network architecture needs appropriate training algorithm. The learning paradigms can be categorized as: supervised, unsupervised and hybrid (reinforcement).

Vohradsky \citep{vohradsky2001} applied ANN to model gene regulation by assuming that the regulatory effect on gene expression of a particular gene can be expressed in the form of ANN. Each neuron in the neural network represents a gene and connectivity between them represents regulatory interactions.  Here each layer of the ANN represents the level of gene expression at time t and output of a neuron at time $t + \Delta t$ can be determined from the expression at time $t$. The advantage the model is that it is continuous, uses a transfer function to transform the inputs to a shape close to those observed in natural processes and does not use artificial elements. Keedwell and his collaborators \citep{keedwell2002} also applied ANN in the purest form for the reconstruction of GRNs from microarray data. The architecture of the neural network was quite simple when dealing with Boolean networks. \citet{hu2006} has proposed a general recurrent neural network (RNN) model for the reverse-engineering of GRNs and to learn their parameters. RNN has been deployed due to its capability to deal with complex temporal behaviour of genetic networks. In this model, time delay between the output of a gene and its effect on another gene has been incorporated. A more recent work by Noman and his colleagues \citep{noman2013} proposed a decoupled-RNN model of GRN. Here, decoupled means dividing the estimation problem of parameters for the complete network into several sub-problems, each of which estimate parameters associated with single gene. This decoupled approach decreases the dimensionality problem and makes the reconstruction of large network feasible. In one of our work \citep{raza2014recurrent}, we also proposed a RNN based hybrid model of GRN that uses extended Kalman filter to estimate and update synaptic weights using Backpropagation Through Time (BPTT) training algorithm.

The ANN approach of GRN inference works well for small size network, i.e., a network of up to 100 genes. This is because of less number of available samples in a Microarray experiment. As the size of the network grows, the number of unknown parameters (interactions) also grows, and that requires a very large number of Microarray samples, which is rarely available in Microarray data.

\subsection{Fuzzy System}
Fuzzy logic is based on the concept of partial truth, i.e., truth values between \emph{completely true} and \emph{completely false}.  For example, using fuzzy logic, propositions can be denoted with degrees of truthfulness and falsehood with the help of a membership function. L.A. Zadeh \citep{zadeh1996fuzzy} was the first who introduced the concept of fuzzy logic to represent vagueness in linguistics and implement and express human knowledge and inference capability in a natural way. In broad sense, fuzzy logic is an extension of multivalue logic. In specific sense, fuzzy logic is a logic system which can be used to model approximate reasoning \citep{cao2006}. Fuzzy logic has been proved to be useful in expert system and other artificial intelligence applications.

A fuzzy system generally consists of three parts:

(i) fuzzy input and output variables, and their fuzzy values,

(ii) fuzzy rules, such as Zadeh-Mamdani's fuzzy rules, Takagi-Sugeno's fuzzy rules, gradual fuzzy rules and recurrent fuzzy rules,

(iii) fuzzy inference methods, which may include fuzzification and defuzzification.

The biological systems behave in a fuzzy manner. Fuzzy logic provides a mathematical framework for modeling and describing biological systems. Literature reports that fuzzy logic has been successfully used for the analysis of microarray data due to its capability to represent non-linear systems, its friendly language to incorporate and edit domain knowledge in the form of fuzzy rules \citep{raza2013soft}. Woolf and Wang \citep{woolf2000} proposed a fuzzy logic based algorithm for analysing gene expression data. The proposed fuzzy model was designed to extract gene triplets (activators, repressors, targets) in yeast gene expression data. The model took ~200 hours to analyse the relationships between 1,898 genes on an 8-processor SGI Origin 2000 system. Later, Ressom and his colleagues \citep{ressom2003} extended and improved the work of Woolf and Wang \citep{woolf2000} in terms of reducing computation time and generalizing the model to accommodate co-activator and co-repressors, in addition to activators, repressors and targets. Reduction in computation time is achieved by applying clustering as a pre-processing step. The improved algorithm achieves a reduction of 50\% computation time. After 3 years, Ram and his colleagues \citep{ram2006} also improved the Woolf and Wang’s fuzzy logic model to predict changes in gene expression values and extracted causal relationship between genes. They have improved searching for activator/repressor regulatory relationship between gene triplets in the microarray data. A pre-processing technique for the fuzzy model has also been proposed to remove redundant data present that makes the model faster. Sun and colleagues \citep{sun2010} applied dynamic fuzzy approach by incorporating structural knowledge to model gene regulatory networks using microarray gene expression data. This technique infers gene interactions in the form of fuzzy rules and able to reveal biological relationships among genes and their products.  The distinguishing feature of this model is that (i) prior structural knowledge on GRN can be incorporated for the purpose of faster convergence of the identification process and (ii) non-linear dynamic property of the GRN can be well captured for the better prediction.

As discussed in previous section, clustering (grouping) Microarray data is also one aspect of analysing it that gives clues about set of co-regulated genes. Fuzzy based clustering algorithm, called fuzzy c-means (FCM) was first introduced by Dunn in 1973 \citep{dunn1973} but implemented by Bezdek \citep{bezdek1981}. The FCM has now become most popular fuzzy clustering algorithm and considered as robust to scale the dataset \citep{wang2008}. A major problem with FCM algorithm for clustering microarray data is the selection of the fuzziness parameter $m$.  The work of \citet{dembele2003} shows that the commonly used value $m = 2$ is not always appropriate. The optimal value for m varies from one dataset to another. They also proposed an empirical method to estimate an adequate value for m, based on the distribution of distances between genes in the given dataset.

In addition to fuzzy-clustering hybrid, fuzzy logic has been hybridized other computational intelligence techniques, including fuzzy and neural network hybrid (called neuro-fuzzy) and fuzzy and genetic algorithm (called fuzzy-genetic). Neuro-fuzzy and fuzzy-genetic hybrid have been successfully applied for GRN inference using Microarray data. The review of the application of neuro-fuzzy  and neuro-genetic for GRN inference can be found in \citep{mitra2011, raza2013soft}.

\subsection{Evolutionary Computation}
Evolutionary computing is a collection of problem-solving techniques based on principles of biological evolution. The functional analogy of evolutionary computing is the natural evolution that relates to a particular kind of problem solving grounded on trial-and-error process. Natural selection means that we have a population of individuals that strive for survival and reproduction. The fitness of these individuals determines how well they succeed in achieving their goals, i.e., presenting their chance for survival and reproduction. Charles Darwin formulated the theory of natural evolution. Over several generations, biological organism evolves according to the principle of natural selection like \emph{survival of the fittest}. The history of evolutionary computing goes back to 1940s. After many decades of research in this area, researchers came up with many evolutionary computing techniques such as evolutionary programming, evolution strategies, genetic algorithm \citep{holland1975} and generic programming \citep{koza1992}.

Genetic algorithms (GAs) are basically optimization techniques inspired by Darwin’s theory of evolution. In fact, it is a search algorithm based on the mechanism of natural selection and survival for the fittest. Here, searching in a population is done from a single point and competitive selection is done in each iterations. The solutions having high “fitness” are recombined with other solutions and then “mutated” by changing the single element of the solution. The purpose of genetic operators, such as crossover and mutation, are to generate new population of solutions for the next generation. Genetic algorithms belong to probabilistic algorithms and are different from random search algorithms because former combines elements of directed and stochastic search. Due to this reason, GAs are found to be more robust than directed search methods. Further, GAs maintain a population of potential solutions; on the other hand, other search techniques process a single point of search space \citep{raza2012evolutionary}.

\citet{noman2005} applied decoupled S-system approach for the inference of effective kinetic parameters from time series gene expression data and applied Trigonometric Differential Evolution (TDE) for the optimization and captures the dynamics of gene expression. Later, \citet{chowdhury2011} extended the work of \citet{noman2005} and applied GA for scoring the networks’ several useful features, such as a Prediction Initialization (PI) algorithm to initialize the individuals, a Flip Operation (FO) for matching values. A refinement algorithm for optimizing sensitivity and specificity of inferred networks was also proposed. Xu and colleagues \citep{xu2009} proposed genetic programming based method for the analysis of microarray datasets, where genetic programming performs classification and feature selection simultaneously. \citet{maulik2011}  studied the performance of three most commonly used computational techniques such as genetic algorithm, simulated annealing and differential evolution for developing fuzzy clusters of gene expression data. Clustering is an unsupervised analysis approach for grouping co-expressed genes together. To improve results of clustering, support vector machine (SVM) has been utilized. A review of application of evolutionary computation for Microarray analysis can be found in \citet{sirbu2010, mitra2011, raza2012evolutionary}.

\subsection{Other AI Based Methods}

Machine learning algorithms, like ANN, are also used to predict interactions between genes of a GRN using Microarray data. But, these algorithms are so complex and work like a black-box. Black-box model means what is happening inside the algorithm is hidden \citep{sirbu2010}. On the other hand, nature-inspired algorithms, in comparison to other algorithms, are simpler in nature and they have been found to be applied in various biological problems from simplest like alignment of sequences to the complex like protein structure prediction \citep{pal2006}. One such type of nature-inspired algorithm is Genetic algorithm which has already been discussed in the previous section.

In the last two decades, several nature-inspired metaheuristic optimization algorithms have been proposed and successfully applied in many optimization problems, including microarray analysis. Fister and colleagues \citep{fister2013} have done a survey of nature-inspired optimization algorithm and listed \~75 nature-inspired algorithms proposed by different researchers, and classified these algorithms into four groups: Swarm intelligence based , Bio-inspired based, Physics-based and Chemistry-based, and Others. algorithms. Ant colony optimization (ACO) is one of the nature-inspired swarm-based optimization algorithm proposed by Marco Dorigo in 1992 \citep{dorigo1992} in his PhD thesis.  ACO is a metaheuristic optimization technique where a set of artificial ants search for optimal solutions in a given optimization problem. Ants use pheromones laid by the other ants as footmarks to follow. Hence, ant reaches the food source by the shortest path using knowledge gained by the other ants. This algorithm can be used for optimization problems, including gene interaction network optimization. ACO has been applied to several bioinformatics problems including sequence alignment, drug designing, 2D protein folding and biological network optimization. \citet{raza2015ant} applied ACO algorithm for inferring highly correlated key gene interactions in a GRN that plays an important role in identifying biomarkers for disease which further helps in drug design. The limitation of proposed algorithm by \citet{raza2015ant} is that it can find out a total number of interactions equal to total number of genes.

PSO has been applied for clustering and feature (genes) selection in microarray data. A k-means clustering based upon PSO has been proposed for microarray data clustering by \citet{deng2005}. The algorithm discovers clusters in microarray data without having any prior knowledge of feasible number of clusters. \citet{chuang2009} applied Binary PSO for feature selection in microarray data.  \citet{sahu2012} also proposed a PSO based feature selection algorithm for cancer microarray data. For the selection of efficient genes from thousands of genes, \citet{chen2014gene} proposed an approach utilizing PSO combined with a decision tree classifier. For the biclustering of microarray data, a comparative study on three nature-inspired algorithms, such as PSO, Shuffled Frog Leaping (SFL) and Cuckoo Search (CS) algorithms, have been done on benchmark gene expression dataset by \citet{balamurugan2014}. The result reports that CS outperforms PSO and SFL for 3 out of 4 datasets. The classification accuracy of simple statistical learning techniques can be enhanced when nature-inspired algorithm are applied for the feature selection. One such study has been carried out by \citet{gunavathi2014}. They performed a comparative analysis of swarm intelligence techniques, such as PSO, cuckoo search (CS), SFL, and SFL with L{\'e}vy flight (SFLLF), for feature selection in cancer classification. The k-nearest neighbour (kNN) classifier is applied to classify the samples. The result shows that k-NN classifier through SFLLF feature selection method outperform PSO, CS, and SFL. Sometimes, DEGs techniques are used for gene selection/filtering or dimension reduction in microarray data where we have a large number of genes (features). The dimension reduction is a preprocessing step whenever we use a machine learning technique for training with gene expression datasets where number of gene are larger than the available samples (generally known as \emph{curse of dimensionality problem}).

Due to advancement in data mining algorithms and tools, it is a keen interest of the researchers to apply these tools to identify patterns of interest in the gene expression data. Association rule mining is one of the most widely used data mining technique that have been applied for gene expression mining by a number of researchers \cite{creighton2003}. Association rules mining may discover biologically useful associations between genes, or between different biological conditions using microarray gene expression data. An association rules are written in the form $A_1 \rightarrow A_2$, where $A_1$ and $A_2$ are disjoint sets of data items. The set $A_2$ is likely to occur whenever the set $A_1$ occurs. Here, the data items may present either highly expressed or repressed genes, or any other facts that state the cellular environment of genes (e.g. diagnosis of a disease samples) \citep{raza2015formal}. Formal Concept Analysis (FCA), introduced by R. Wille in early 1980s, is another data mining technique based on lattice theory. It has been widely used for the analysis of binary relational data. Like other computational techniques, FCA has also been applied in microarray analysis, gene expression mining, gene expression clustering, finding genes in gene regulatory networks, and so on. A review FAC for the analysis and knowledge discovery from gene expression and other biological data can be found in \citet{raza2015formal}.

\section{Conclusions, Discussions and Future Challenges}
The bioinformatics is an interdisciplinary area of study where one of the objectives is to deal with the analysis and interpretation of large sets of data generated from various large-scale biological experiments, including Microarrays.  Microarray technology is one of the powerful tools used to measure genome wide expression levels of genes. As microarrays technologies have become more prevalent, the challenges associated with collecting, managing, and analyzing the data from each experiment have essentially increased. With the help of these technologies, researchers can find out answer of some challenging questions like: (i) what are the functions of different genes? (ii) In what cellular processes do they participate? (iii) how genes are regulated? (iv) how genes and its products (proteins) do interact, and what are these interaction networks? (v) how expression level of genes differs in different cell types and states? (vi) how expressions of genes are affected by various disease or drug treatments?

In this chapter, four broad categories of problems have been tackled for the analysis of Microarrays:
\begin{enumerate}[(i)]
  \item \emph{Identification of differentially expressed genes:} It helps us in the selection of few relevant genes and elimination of irrelevant genes for further study. It also solves the dimensionality problem of machine learning techniques by filtering differentially expressed genes over various samples and training a classifier with the selected number of genes only.
  \item \emph{Cancer classification using gene expression:} Classification of patient based on gene expression profile is another important issue for the analysis of microarray data. The application of AI-based techniques for cancer classification based microarray data has been discussed.
  \item \emph{Clustering genes, conditions or both:} Another important aspect of analyzing microarray data is finding natural groupings among genes, which can be done using clustering techniques. Clustering is an unsupervised learning technique that plays a vital role in providing a “class label” to unlabeled data and it can be used to identify set of co-regulated genes.
  \item \emph{Inferring gene interaction network:} The gene interaction network plays an important role in identifying root-cause of various diseases. Inferring gene interaction network from gene expression profiles is one of other aspect of analyzing microarray data. The application of AI-based techniques for GRN inference has been covered in length and various resources for further study has been listed.
\end{enumerate}

\subsection*{Future Challenges}
Microarray technology is a high-throughput experimental approach that measures the genome-wide expression of genes and data are produced in large-scale. Hence, analysing these data to infer useful information is big challenge. Some of the future directions for the analysis of microarray data are as follows:

\begin{enumerate}[(i)]
  \item One of the main drawbacks of microarray technology is that data generated by these experiments contain noises and are not much reliable. Hence, before the data is analysed, we must apply sophisticated noise removal and data normalization technique.
  \item Application of machine learning techniques in genome-wide analysis of microarray data creates the problem of dimensionality. Hence, some techniques are required to identify differentially expressed genes (DEGs). Statistical techniques, such as fold change, t-test, ANOVA, etc. dominates in the identification of DEGs. Hence, it is needed to explore the application of computational intelligence to tackle the problem of DEGs.
  \item Another biggest challenge is to develop a single classifier which is best suitable for classification of all types of cancer gene expression data into meaningful number of classes. Nature inspired optimization techniques such as Ant Colony Optimization (ACO) \citep{dorigo1992}, Artificial Bee Colony optimization (ABC) \cite{karaboga2005}, Cuckoo Search \cite{yangdeb2009}, Particle Swarm Optimization (PSO) \cite{kennedy1995}, Spider Monkey Algorithm \citep{bansal2014} and so on, are successfully being used in many challenging problems. In the future work, one can hybridize these nature inspired optimization techniques with different classifiers for better classification accuracy.
  \item For the gene clustering problem, one can apply fuzzy based clustering techniques (such as Fuzzy C-Means) to group genes or patient or both. Even, ranked based classification techniques can be applied.
  \item Inference of gene interaction networks using gene expression profile is another open area where computational intelligence can be applied to identify interactions among given set of genes. Hybrid algorithms (for example, fusion of neural networks, genetic algorithms and/or fuzzy logic and other nature-inspired algorithms) can be applied for the said purpose.
\end{enumerate}

\section*{Acknowledgements}
The author acknowledges the funding received from University Grants Commission, Govt. of India through research grant 42-1019/2013(SR).
\bibliographystyle{apalike}
\bibliography{bibfile}

\end{document}